\pdfoutput=1
\documentclass[11pt]{article}

\usepackage[final]{acl}
\usepackage{acl}
\usepackage{booktabs}
\usepackage{times}
\usepackage{latexsym}
\usepackage{amsmath}
\usepackage[T1]{fontenc}
\usepackage[utf8]{inputenc}
\usepackage{subcaption}
\usepackage[inline]{enumitem}

\usepackage{microtype}

\usepackage{inconsolata}

\usepackage{graphicx}
\usepackage{longtable}
\usepackage{booktabs}
\usepackage{array}
\usepackage{geometry}
\usepackage{fancyvrb}
\usepackage{xcolor}

\author{Sabri Boughorbel \hspace{11mm} Fahim Dalvi \hspace{11mm} Nadir Durrani \hspace{11mm} Majd Hawasly \\
Qatar Computing Research Institute, HBKU, Doha, Qatar \\ 
{\tt \{sboughorbel,faimaduddin,ndurrani,mhawasly\}@hbku.edu.qa} \\ 
}

\title{Beyond the Leaderboard: Understanding Performance Disparities in \\Large Language Models via Model Diffing}

\begin{document}

\maketitle

\begin{abstract}
As fine-tuning becomes the dominant paradigm for improving large language models (LLMs), understanding what changes during this process is increasingly important. Traditional benchmarking often fails to explain \textit{why} one model outperforms another. In this work, we  
use \textbf{model diffing}, a mechanistic interpretability approach, to analyze the specific capability differences between \textbf{Gemma-2-9b-it} and a \textbf{SimPO-enhanced} variant. Using \textbf{crosscoders}, we identify and categorize latent representations that 
differentiate the two models. We find 
that SimPO acquired latent concepts predominantly enhance safety mechanisms (+32.8\%), multilingual capabilities (+43.8\%), and instruction-following (+151.7\%), while its additional training also reduces emphasis on model self-reference (-44.1\%) and hallucination management (-68.5\%). Our analysis shows that model diffing can yield fine-grained insights beyond leaderboard metrics, attributing performance gaps to concrete mechanistic capabilities. This approach offers a transparent and targeted framework for comparing LLMs.
\end{abstract}

\section{Introduction}

Open-weight large language models (LLMs) have transformed the AI landscape, making it increasingly challenging for academic and low-resource organizations to train competitive models from scratch~\cite{yang2025qwen3,team2025gemma, fanarteam2025,liu2024deepseek, grattafiori2024llama}. Instead, fine-tuning models has become the mainstream approach for developing new capabilities and improving performance. Understanding precisely \textit{what} changes during fine-tuning and \textit{why} certain models outperform others remains challenging.

Current evaluation paradigms rely heavily on benchmarks, which, while useful for capturing specific aspects of model performance, come with significant limitations. 
As benchmarks gain popularity, the risk of data contamination increases~\cite{xu2024benchmark}, and over time, they can become saturated, making them costly to update or replace. Moreover, benchmarks are susceptible to gaming~\cite{metascandal}, which undermines their reliability. Additionally, a benchmark captures only a specific aspect of performance, as evidenced in the continuous development and release of new benchmarks targeting different domains and skills. As such, benchmarking highlights certain dimensions of difference but may overlook many others.
On the other hand, human evaluations, such as LMArena~\cite{chiang2024chatbotarenaopenplatform}, offer more authentic and wide-ranging assessment, but they are resource-intensive and can still be swayed by superficial factors like response style and verbosity rather than true differences in model capability~\cite{lmarenastyle,singh2025leaderboard}.

Another approach to uncovering what a model encodes is structural probing~\cite{belinkov-etal-2017-neural, hewitt-manning-2019-structural, kantamneni2025are}, which involves training small classifiers (\textit{probes}) to predict specific properties from the model's internal representations. This approach has, for example, been applied to study how transfer learning impacts linguistic knowledge in deep NLP models~\cite{durrani-etal-2021-transfer}. However, it may lack the sensitivity needed to detect subtle differences between closely related models and typically requires prior knowledge of the properties being investigated.

In this paper, we use Model Diffing \citep{lindsey2024sparse, minder2025robustly} with crosscoders to analyze the latent representations of two models.
For a use case, we investigate the improvements brought by the \textit{Simplified Preference Optimization (SimPO)} technique~\cite{meng2024simposimplepreferenceoptimization} which has been promoted as a significant advancement in RLHF, credited with boosting the performance of \texttt{Gemma-2-9b-it}~\cite{team2024gemma} across both leaderboard benchmark scores and human preference evaluations. However, a closer look reveals that these improvements may be largely attributable to superficial factors such as stylistic polishing and output formatting rather than genuine gains in reasoning, factual accuracy, or task competence; see Table~\ref{tab:class-changes2}. This raises a critical question: \textit{Are fine-tuning methods like SimPO truly enhancing model capabilities, or merely optimizing for appearances that game existing evaluation setups?}

\begin{table}[!]
\small
\centering
\begin{tabular}{l|ccc}
\toprule
Category & $\Delta^{ELO}$ & $\Delta_{\text{style}}^{ELO}$ & Diff \\
\midrule
Math & 10	& -12	& -22 \\
Chinese & 40 & 27 & -13 \\

Coding & 19&	6&	-13 \\

Russian & 25 &  15 & -10  \\
Hard Prompts & 25 &	17	&-8 \\
Multi-Turn & 27	& 20	& -7 \\
German & 34 & 27 & -7 \\
Creative Writing & 33	& 26	& -7 \\
Overall & 20	&15 	&-5 \\
Instruction Following & 12	& 8	& -4 \\
English & 23 & 20&  -3\\

\bottomrule
\end{tabular}
\caption{The delta of LMArena Elo scores,\protect\footnotemark\;  per category for \texttt{gemma-2-9b-it-SimPO} compared to \texttt{gemma-2-9b-it}, without ($\Delta^{ELO}$) and with style control ($\Delta_{\text{style}}^{ELO}$) applied. The differences between the deltas (Diff) indicate that style alone accounts for a considerable portion of the observed improvements of SimPO. The full Elo results are reported in Table~\ref{tab:class-performance-default}.}
\label{tab:class-changes2}
\end{table}

\footnotetext{https://lmarena.ai/leaderboard/text}

We apply \textbf{Model Diffing} to analyze the latent representations of \texttt{Gemma-2-9b-it} and its fine-tuned variant \texttt{Gemma-2-9b-it-SimPO}. By additionally contrasting both with their shared base model (\texttt{Gemma-2-9b-pt}), we identify and categorize representation-level changes that help explain observed performance differences. This mechanistic approach provides a nuanced view of how SimPO fine-tuning alters model behavior, revealing both gains and potential regressions in capabilities.

Our analysis shows that SimPO fine-tuning leads to targeted shifts in model capabilities rather than uniform improvements. We find substantial increases in \textbf{safety and moderation} (\textbf{+32.8\%}), \textbf{multilingual and stylistic processing} (\textbf{+43.8\%}), and \textbf{instruction-following} (\textbf{+151.7\%}), aligning with SimPO's optimization for alignment and human preference signals. At the same time, we observe notable regressions in \textbf{hallucination detection} (\textbf{--68.5\%}), \textbf{model self-reference} (\textbf{--44.1\%}), and \textbf{structured output generation} (\textbf{--37.1\%}), suggesting a trade-off between confident, polished outputs and internal verification or reasoning. These changes point to a broader shift: SimPO appears to prioritize \textit{fluency and alignment cues} over deliberation or factual introspection, which may partially explain its improved preference ratings despite mixed technical performance. Crucially, these shifts are only visible through model diffing, not from benchmark scores or leaderboard deltas, highlighting the need for deeper mechanistic diagnostics in evaluating LLM enhancements.

Thus, the contribution of this work lies in presenting a generalizable and interpretable
methodology for isolating and categorizing behavioral changes between closely related models using
latent-space diffing via crosscoders. We use  SimPO as a case study to showcase the methodology and the behavioral taxonomy that emerges from it, but this method is general and could apply to other training approaches. See Appendix~\ref{appx:dpo} for additional results with regard to Direct Preference Optimization (DPO) fine-tuning~\cite{rafailov2023direct}.

\section{Methodology}

To analyze model differences, we employ the recently developed technique of \textbf{Model Diffing} using \textbf{crosscoders}~\cite{lindsey2024sparse, minder2025robustly}. Crosscoders are a specialized form of sparse autoencoders~\cite{yun2021transformer, bricken2023monosemanticity, huben2023sparse} that learn a shared dictionary of interpretable latent concepts across two models. This enables us to identify how internal representations shift or diverge after fine-tuning.

\subsection{Model Diffing with Crosscoders}

The crosscoder workflow involves three main steps:\begin{enumerate*}[label=(\arabic*)]
    \item A shared dictionary is trained to reconstruct the activation patterns from both models.
    \item For each latent dimension, a pair of decoder directions is learned, one for each model.
    \item The differences between these directions are analyzed to identify model-specific capabilities.
\end{enumerate*}

By comparing the norm differences between corresponding latent vectors in each model, we can identify concepts that are uniquely important to one model relative to the other. The norm difference between two models \( M_1 \) and \( M_2 \) is defined as:
$$\Delta_{\text{norm}}(j) = \frac{1}{2} \left( \frac{\|\mathbf{d}_j^{\text{M}_2}\|_2 - \|\mathbf{d}_j^{\text{M}_1}\|_2}{\max(\|\mathbf{d}_j^{\text{M}_2}\|_2, \|\mathbf{d}_j^{\text{M}_1}\|_2)} + 1 \right)$$
where \(\mathbf{d}_j^{M_1}\) and \(\mathbf{d}_j^{M_2}\) are the decoder vectors corresponding to latent \(j\) in the two models.

This approach may suffer from two known failure modes: \textit{Complete Shrinkage} and \textit{Latent Decoupling}, which can cause shared latents to be misclassified as model-specific. To mitigate this, we apply the \textit{Latent Scaling} technique~\cite{minder2025robustly, wright2024addressing}, which estimates two coefficients, \(\nu^\epsilon\) and \(\nu^r\), to more accurately measure latent presence across models. Combined with \textit{BatchTopK} training~\cite{bussmann2024batchtopk, GaoTTGTRSL025}, this enables identification of latents that are causally unique to the fine-tuned or base model.

\subsection{Experimental Setup}
\label{subsection:experiment_setup}

We trained crosscoders to study activation patterns across three variants of the Gemma-2-9b model~\footnote{The trained crosscoder models and data are released at https://github.com/bsabri/LLMDiff}. Specifically, we employed the BatchTopK Sparse Autoencoder (SAE) training method with a latent dimensionality of 114,688,  top-$k = 100$ and learning rate of 1e-4. BatchTopK has been shown to outperform the traditional $L_1$-based crosscoder training loss~\cite{bussmann2024batchtopk,GaoTTGTRSL025}. Following prior work~\cite{lieberum2024gemma}, we selected layer 20 for analysis. Crosscoders were trained using 200M tokens from a mixed corpus comprising the FineWeb~\cite{penedo24fineweb} and LMSys datasets~\cite{zheng2023lmsys}. We considered the following open model variants:
\begin{enumerate}[label=(\arabic*)]
    \item \textbf{Gemma-2-9b-pt:}\footnote{https://hf.co/google/gemma-2-9b} The original pretrained model from Google.
    \item \textbf{Gemma-2-9b-it:\footnote{https://hf.co/google/gemma-2-9b-it}} The instruction-tuned model with supervised fine-tuning and alignment.
   \item \textbf{Gemma-2-9b-it-SimPO:\footnote{https://hf.co/princeton-nlp/gemma-2-9b-it-SimPO}} The SimPO-enhanced variant of the instruction-tuned model.
\end{enumerate}

\begin{figure*}[t]
     \centering
     \begin{subfigure}[b]{0.30\textwidth}
         \centering
         \includegraphics[width=\linewidth]{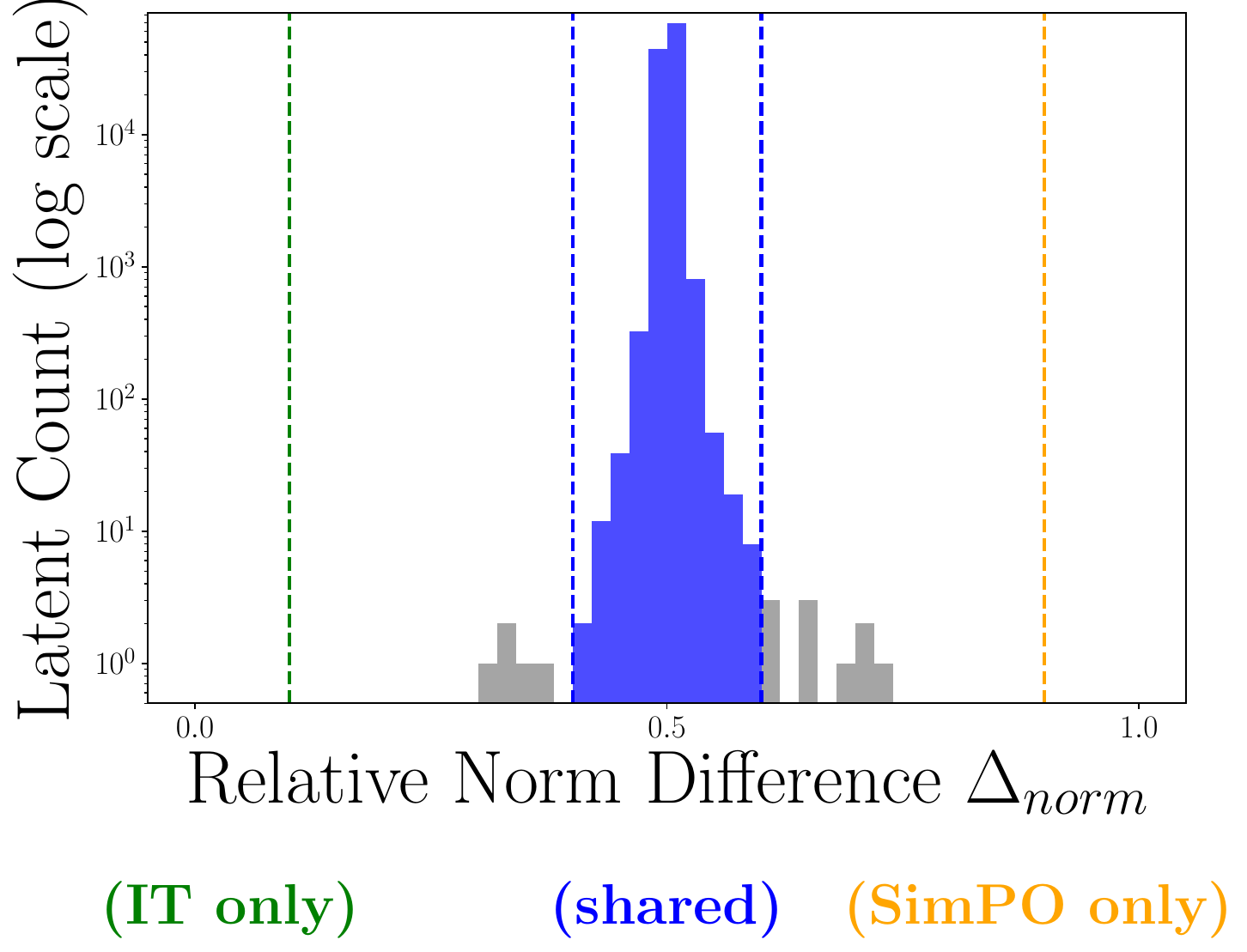}
         \caption{Latent distribution of $\Delta_{\text{norm}}$ for \texttt{SimPO-it} crosscoder.}
         \label{fig:a}
     \end{subfigure}
     \hfill
     \begin{subfigure}[b]{0.30\textwidth}
         \centering
         \includegraphics[width=\textwidth]{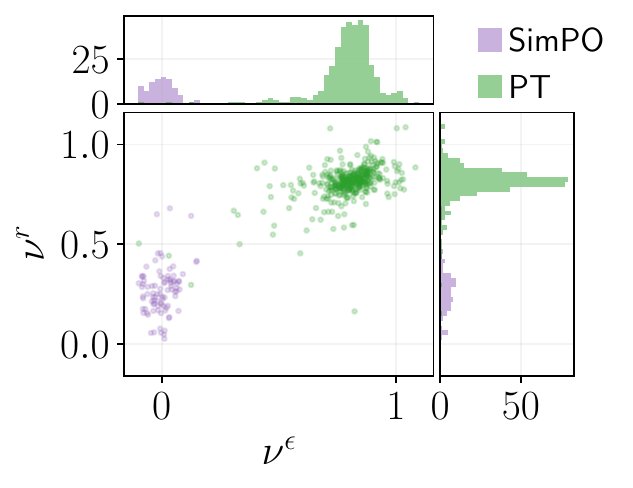}
        \caption{\texttt{pt-SimPO} crosscoder latents.}
         \label{fig:b}
     \end{subfigure}
     \hfill
     \begin{subfigure}[b]{0.30\textwidth}
         \centering
        \includegraphics[width=\textwidth]{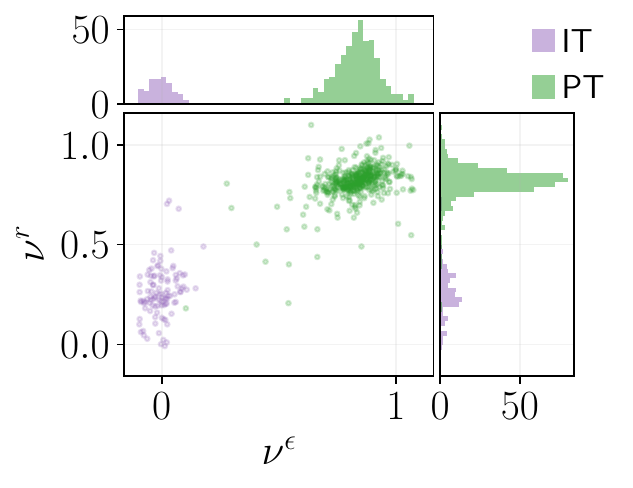}
        \caption{\texttt{pt-it} crosscoder latents.}
         \label{fig:c}
     \end{subfigure}
        \caption{Identification of distinct latents across trained crosscoders. Figure~\ref{fig:a} shows that no latents are unique to \texttt{SimPO} or \texttt{it} models for $\Delta_{norm}$ thresholds  $<0.1$ or  $>0.9$. Figure~\ref{fig:b} and \ref{fig:c} show the distribution of latents w.r.t. $\nu^\epsilon$ and $\nu^r$ coefficient in Latent Scaling method. The purple latents are, respectively, unique to \texttt{SimPO} and \texttt{it} in crosscoders \texttt{pt-SimPO} and \texttt{pt-it}. We identified 92 latents unique to \texttt{SimPO} in \texttt{pt-SimPO}  and 113 latents unique to \texttt{it} in \texttt{it-pt} crosscoder. These latents are taken further in downstream analysis as described in Section~\ref{subsection:experiment_setup}.  }
        \label{fig:three graphs}
\end{figure*}

\subsection{Design Iteration}

The intuitive starting point for our analysis was to compare \texttt{Gemma-2-9b-it} (instruction-tuned) with \texttt{Gemma-2-9b-it-SimPO} (SimPO-enhanced), since our primary interest was in understanding what makes SimPO effective. However, this direct comparison showed nonsignificant difference. The distribution of the latent normal difference showed concepts falling into a generic ``other'' category as illustrated in Figure~\ref{fig:a} with a norm difference mostly in the range of 0.3 to 0.6, which were too subtle to yield meaningful interpretability.

This outcome led us to reassess our approach. A likely explanation is that once a model has undergone instruction tuning, further improvements like SimPO operate within a narrow behavioral subspace. They may modify surface-level generation preferences or alignment signals rather than introducing fundamentally new internal representations. As a result, SimPO's changes are less visible at the level of latent activation dynamics. Since we used BatchTopK method for training the crosscoder, only causally distinct latents are retained~\cite{minder2025robustly}.

Thus, to better capture and interpret meaningful representational shifts, we revised our experimental setup to compare each fine-tuned model (\texttt{it} and \texttt{it-SimPO}) directly with the shared base model (\texttt{Gemma-2-9b-pt}). This change provided a clearer view of how instruction tuning and subsequent enhancements shape the model's internal structure, allowing us to trace the emergence and transformation of capability-related latents more effectively.  We identified 92 latents unique to \texttt{SimPO} in \texttt{pt-SimPO}  and 113 latents unique to \texttt{it} in \texttt{it-pt} crosscoder. See Figure~\ref{fig:b} and \ref{fig:c}.

We extracted documents that strongly activated the identified latents from the training dataset, and used a large language model (Claude 3 Opus -- \texttt{anthropic/claude-3-opus-20240229}) to annotate and categorize these documents~\cite{mousi-etal-2023-llms,karvonen2025saebench, paulo2024automatically}. The result was a taxonomy of 30 capability categories grouped under 7 major classes (see Appendix~\ref{appx:annotation} for the annotation pipeline, and Section~\ref{appx:classes} for the full taxonomy). We measured the normalized frequency of each category for the identified latents in the two studied models, and we analyze the differences of these frequencies as a measure to understand the source of performance disparities.

To move beyond aggregate performance metrics, we conduct a representational analysis of latent concept shifts introduced by SimPO fine-tuning. Our comparison addresses two key research questions: \textbf{(i)}~\textit{Which latent capabilities are strengthened through SimPO's preference-driven fine-tuning?} and \textbf{(ii)}~\textit{What capabilities are diminished or deprioritized as a result of this optimization?}

Table~\ref{tab:performance-comparison} summarizes the distribution of frequencies of latent categories across the seven high-level classes, highlighting the most significant shifts in model behavior. We report and interpret these changes next. The most frequent fine-grained categories for the two models can be found in Appendix~\ref{sec:detailed-finegrained-results}.

\begin{table}[!]
\small
\centering
\begin{tabular}{p{2.7cm}|p{0.5cm}p{0.7cm}p{0.7cm}p{1.0cm}}
\toprule
Class & \texttt{it} & \texttt{SimPO} & Diff & Change \\
\midrule
Linguistic Capabilities & 6.25 & 8.99 & +2.74 & +43.8\% \\
Safety \& Content Moderation & 16.07 & 21.35 & +5.28 & +32.8\% \\
Information Processing & 16.96 & 17.98 & +1.01 & +6.0\% \\
Format \& Structure Control & 10.71 & 11.24 & +0.52 & +4.9\% \\
User Interaction Management & 14.29 & 12.36 & -1.93 & -13.5\% \\
Specialized Capabilities & 28.57 & 23.60 & -4.98 & -17.4\% \\
Error Handling \& Quality Control & 6.25 & 4.49 & -1.76 & -28.1\% \\
\bottomrule
\end{tabular}
\caption{Class-level latent count changes (\%) between \texttt{Gemma-2-9b-it} and \texttt{Gemma-2-9b-it-SimPO} models. Fine-grained results are available in Appendix~\ref{sec:detailed-finegrained-results} }

\label{tab:performance-comparison}
\end{table}
\section{Findings}

\subsection{Enhanced Capabilities in SimPO}

Among the latent concepts identified through model diffing, a notable subset becomes more prominent in the SimPO-enhanced model. These capabilities largely align with the goals of preference optimization, such as improving stylistic fluency, safety, and adherence to user instructions. Below, we summarize the most significant gains across categories like alignment, multilingual processing, and factual verification.

\paragraph{Linguistic capabilities (+43.8\%)} SimPO demonstrates enhanced multilingual capabilities. This explains the observed  improvements in English, German and Chinese. However low-resource languages (Japanese, Korean) show regression on LMArena score and that was not captured by the latents possibly due to the lack of such data in the crosscoder training data.

\paragraph{Safety \& Content Moderation (+32.8\%)} The most increase occurred in safety mechanisms, with Sexual Content Filtering showing the largest growth. Other notable increases include Minor Protection and Stereotype \& Bias Detection. This suggests that SimPO prioritizes alignment with human values and safety guidelines.

\subsection{Diminished Capabilities in SimPO}

While SimPO strengthens many alignment-related capabilities, our analysis also reveals a set of latent concepts that decrease in prominence. These diminished features point to potential trade-offs, including reduced introspection, hallucination detection, and structured reasoning. We highlight the most notable regressions and discuss their implications for model reliability and robustness:

\paragraph{Code Generation \& Math }Technical capabilities (e.g., related to code and math) show decrease of -17.4\%, potentially indicating a shift toward general-purpose conversational abilities. This is reflected in the LMArena score.

\section{Discussion}

Our mechanistic analysis of differences between \texttt{Gemma-2-9b-it} and its SimPO-enhanced variant reveals that SimPO's performance improvements stem from specific capability shifts rather than uniform enhancements. The most significant changes align with the intended goals of preference optimization: improving safety, alignment with human preferences, and following instructions precisely.

The substantial increases in safety mechanisms (+32.8\%) and instruction-following capabilities suggest that SimPO effectively incorporates human preferences regarding appropriate content and response formats. The dramatic enhancement in template and instruction following (+151.7\%) explains why SimPO often produces more aesthetically pleasing and well-structured responses.

However, our analysis also reveals trade-offs. The decreased emphasis on hallucination detection (–68.5\%) raises questions about whether SimPO sacrifices some self-monitoring capabilities in favor of producing more confident-sounding responses. Similarly, the reduction in query classification suggests that SimPO may take a more direct approach to generating responses than first analyzing the query type.

These findings help explain the mixed results observed in different evaluation contexts. In benchmark tests that reward accurate, well-formatted responses, SimPO's enhanced instruction-following and factual verification capabilities provide an advantage. In open-ended evaluations like lm-arena, human evaluators may be influenced by SimPO's improved stylistic qualities and reduced self-reference, even if some technical capabilities show modest decreases.

\section{Conclusion}

This work demonstrates that model diffing via crosscoders offers valuable insights beyond traditional benchmark evaluations. By mechanistically analyzing the latent representations that distinguish Gemma-2-9b-it from its SimPO variant, we reveal that performance differences stem from specific capability shifts rather than uniform improvements. 

Our findings highlight how SimPO substantially enhances safety mechanisms, instruction-following capabilities, and multilingual processing while reducing emphasis on model self-reference and certain technical capabilities. These insights help explain both the strengths and limitations observed in different evaluation contexts.

Our work suggests that the field should move beyond leaderboard comparisons toward more nuanced analyses of what specifically changes when models are fine-tuned. Model diffing provides a promising framework for understanding performance disparities in terms of specific capabilities rather than opaque metrics, enabling more transparent and meaningful evaluations of LLM enhancements.

\section*{Limitations}
Our study has several limitations. First, we focused on two model pairs (Gemma-2-9b-it and its SimPO and DPO variants), and the patterns we observed might not generalize to other models or fine-tuning approaches. Second, while our crosscoder-based analysis provides insights into capability differences, it cannot definitively establish causal relationships between these differences and specific performance outcomes.

\section*{Ethical considerations}
We use LLM according to their intended use, and we used academic-purpose code that is shared for research objectives. AI tools were used to rephrase and improve exposition of sections of the paper. 
\bibliography{custom}

\begin{thebibliography}{31}
\providecommand{\natexlab}[1]{#1}

\bibitem[{Belinkov et~al.(2017)Belinkov, Durrani, Dalvi, Sajjad, and Glass}]{belinkov-etal-2017-neural}
Yonatan Belinkov, Nadir Durrani, Fahim Dalvi, Hassan Sajjad, and James Glass. 2017.
\newblock \href {https://doi.org/10.18653/v1/P17-1080} {What do neural machine translation models learn about morphology?}
\newblock In \emph{Proceedings of the 55th Annual Meeting of the Association for Computational Linguistics (Volume 1: Long Papers)}, pages 861--872, Vancouver, Canada. Association for Computational Linguistics.

\bibitem[{Bricken et~al.(2023)Bricken, Templeton, Batson, Chen, Jermyn, Conerly, Turner, Anil, Denison, Askell, Lasenby, Wu, Kravec, Schiefer, Maxwell, Joseph, Hatfield-Dodds, Tamkin, Nguyen, McLean, Burke, Hume, Carter, Henighan, and Olah}]{bricken2023monosemanticity}
Trenton Bricken, Adly Templeton, Joshua Batson, Brian Chen, Adam Jermyn, Tom Conerly, Nick Turner, Cem Anil, Carson Denison, Amanda Askell, Robert Lasenby, Yifan Wu, Shauna Kravec, Nicholas Schiefer, Tim Maxwell, Nicholas Joseph, Zac Hatfield-Dodds, Alex Tamkin, Karina Nguyen, and 6 others. 2023.
\newblock Towards monosemanticity: Decomposing language models with dictionary learning.
\newblock \emph{Transformer Circuits Thread}.
\newblock Https://transformer-circuits.pub/2023/monosemantic-features/index.html.

\bibitem[{Bussmann et~al.(2024)Bussmann, Leask, and Nanda}]{bussmann2024batchtopk}
Bart Bussmann, Patrick Leask, and Neel Nanda. 2024.
\newblock Batchtopk sparse autoencoders.
\newblock In \emph{NeurIPS 2024 Workshop on Scientific Methods for Understanding Deep Learning}.

\bibitem[{Chiang et~al.(2024)Chiang, Zheng, Sheng, Angelopoulos, Li, Li, Zhu, Zhang, Jordan, Gonzalez, and Stoica}]{chiang2024chatbotarenaopenplatform}
Wei-Lin Chiang, Lianmin Zheng, Ying Sheng, Anastasios~Nikolas Angelopoulos, Tianle Li, Dacheng Li, Banghua Zhu, Hao Zhang, Michael Jordan, Joseph~E. Gonzalez, and Ion Stoica. 2024.
\newblock \href {https://proceedings.mlr.press/v235/chiang24b.html} {Chatbot arena: An open platform for evaluating {LLM}s by human preference}.
\newblock \emph{Proceedings of the 41st International Conference on Machine Learning}, 235:8359--8388.

\bibitem[{Durrani et~al.(2021)Durrani, Sajjad, and Dalvi}]{durrani-etal-2021-transfer}
Nadir Durrani, Hassan Sajjad, and Fahim Dalvi. 2021.
\newblock \href {https://doi.org/10.18653/v1/2021.findings-acl.438} {How transfer learning impacts linguistic knowledge in deep {NLP} models?}
\newblock In \emph{Findings of the Association for Computational Linguistics: ACL-IJCNLP 2021}, pages 4947--4957, Online. Association for Computational Linguistics.

\bibitem[{Fanar-Team et~al.(2025)Fanar-Team, Abbas, Ahmad, Alam, Altinisik, Asgari, Boshmaf, Boughorbel, Chawla, Chowdhury, Dalvi, Darwish, Durrani, Elfeky, Elmagarmid, Eltabakh, Fatehkia, Fragkopoulos, Hasanain, Hawasly, Husaini, Jung, Lucas, Magdy, Messaoud, Mohamed, Mohiuddin, Mousi, Mubarak, Musleh, Naeem, Ouzzani, Popovic, Sadeghi, Sencar, Shinoy, Sinan, Zhang, Ali, Kheir, Ma, and Ruan}]{fanarteam2025}
Fanar-Team, Ummar Abbas, Mohammad~Shahmeer Ahmad, Firoj Alam, Enes Altinisik, Ehsannedin Asgari, Yazan Boshmaf, Sabri Boughorbel, Sanjay Chawla, Shammur Chowdhury, Fahim Dalvi, Kareem Darwish, Nadir Durrani, Mohamed Elfeky, Ahmed Elmagarmid, Mohamed Eltabakh, Masoomali Fatehkia, Anastasios Fragkopoulos, Maram Hasanain, and 23 others. 2025.
\newblock \href {https://arxiv.org/abs/2501.13944} {Fanar: An arabic-centric multimodal generative ai platform}.
\newblock \emph{Preprint}, arXiv:2501.13944.

\bibitem[{Gao et~al.(2025)Gao, la~Tour, Tillman, Goh, Troll, Radford, Sutskever, Leike, and Wu}]{GaoTTGTRSL025}
Leo Gao, Tom~Dupr{\'{e}} la~Tour, Henk Tillman, Gabriel Goh, Rajan Troll, Alec Radford, Ilya Sutskever, Jan Leike, and Jeffrey Wu. 2025.
\newblock \href {https://openreview.net/forum?id=tcsZt9ZNKD} {Scaling and evaluating sparse autoencoders}.
\newblock In \emph{The Thirteenth International Conference on Learning Representations, {ICLR} 2025, Singapore, April 24-28, 2025}. OpenReview.net.

\bibitem[{Gemma~Team et~al.(2024)Gemma~Team, Mesnard, Hardin, Dadashi, Bhupatiraju, Pathak, Sifre, Rivi{\`e}re, Kale, Love et~al.}]{team2024gemma}
{}~Gemma~Team, Thomas Mesnard, Cassidy Hardin, Robert Dadashi, Surya Bhupatiraju, Shreya Pathak, Laurent Sifre, Morgane Rivi{\`e}re, Mihir~Sanjay Kale, Juliette Love, and 1 others. 2024.
\newblock Gemma: Open models based on gemini research and technology.
\newblock \emph{arXiv preprint arXiv:2403.08295}.

\bibitem[{Grattafiori et~al.(2024)Grattafiori, Dubey, Jauhri, Pandey, Kadian, Al-Dahle, Letman, Mathur, Schelten, Vaughan et~al.}]{grattafiori2024llama}
Aaron Grattafiori, Abhimanyu Dubey, Abhinav Jauhri, Abhinav Pandey, Abhishek Kadian, Ahmad Al-Dahle, Aiesha Letman, Akhil Mathur, Alan Schelten, Alex Vaughan, and 1 others. 2024.
\newblock The llama 3 herd of models.
\newblock \emph{arXiv preprint arXiv:2407.21783}.

\bibitem[{Hewitt and Manning(2019)}]{hewitt-manning-2019-structural}
John Hewitt and Christopher~D. Manning. 2019.
\newblock \href {https://doi.org/10.18653/v1/N19-1419} {{A} structural probe for finding syntax in word representations}.
\newblock In \emph{Proceedings of the 2019 Conference of the North {A}merican Chapter of the Association for Computational Linguistics: Human Language Technologies, Volume 1 (Long and Short Papers)}, pages 4129--4138, Minneapolis, Minnesota. Association for Computational Linguistics.

\bibitem[{Huben et~al.(2023)Huben, Cunningham, Smith, Ewart, and Sharkey}]{huben2023sparse}
Robert Huben, Hoagy Cunningham, Logan~Riggs Smith, Aidan Ewart, and Lee Sharkey. 2023.
\newblock Sparse autoencoders find highly interpretable features in language models.
\newblock In \emph{The Twelfth International Conference on Learning Representations}.

\bibitem[{Kantamneni et~al.(2025)Kantamneni, Engels, Rajamanoharan, Tegmark, and Nanda}]{kantamneni2025are}
Subhash Kantamneni, Joshua Engels, Senthooran Rajamanoharan, Max Tegmark, and Neel Nanda. 2025.
\newblock \href {https://openreview.net/forum?id=rNfzT8YkgO} {Are sparse autoencoders useful? a case study in sparse probing}.
\newblock In \emph{Forty-second International Conference on Machine Learning}.

\bibitem[{Karvonen et~al.(2025)Karvonen, Rager, Lin, Tigges, Bloom, Chanin, Lau, Farrell, McDougall, Ayonrinde et~al.}]{karvonen2025saebench}
Adam Karvonen, Can Rager, Johnny Lin, Curt Tigges, Joseph Bloom, David Chanin, Yeu-Tong Lau, Eoin Farrell, Callum McDougall, Kola Ayonrinde, and 1 others. 2025.
\newblock Saebench: A comprehensive benchmark for sparse autoencoders in language model interpretability.
\newblock \emph{arXiv preprint arXiv:2503.09532}.

\bibitem[{Li et~al.(2024)Li, Angelopoulos, and Chiang}]{lmarenastyle}
Tianle Li, Anastasios Angelopoulos, and Wei-Lin Chiang. 2024.
\newblock \href {https://blog.lmarena.ai/blog/2024/style-control/} {Does style matter? disentangling style and substance in chatbot arena}.
\newblock Accessed: 2025-05-19.

\bibitem[{Lieberum et~al.(2024)Lieberum, Rajamanoharan, Conmy, Smith, Sonnerat, Varma, Kram{\'a}r, Dragan, Shah, and Nanda}]{lieberum2024gemma}
Tom Lieberum, Senthooran Rajamanoharan, Arthur Conmy, Lewis Smith, Nicolas Sonnerat, Vikrant Varma, J{\'a}nos Kram{\'a}r, Anca Dragan, Rohin Shah, and Neel Nanda. 2024.
\newblock Gemma scope: Open sparse autoencoders everywhere all at once on gemma 2.
\newblock \emph{arXiv preprint arXiv:2408.05147}.

\bibitem[{Lindsey et~al.(2024)Lindsey, Templeton, Marcus, Conerly, Batson, and Olah}]{lindsey2024sparse}
Jack Lindsey, Adly Templeton, Jonathan Marcus, Thomas Conerly, Joshua Batson, and Christopher Olah. 2024.
\newblock \href {https://transformer-circuits.pub/2024/crosscoders/index.html} {Sparse crosscoders for cross-layer features and model diffing}.
\newblock \emph{Transformer Circuits Thread}.
\newblock Research Update.

\bibitem[{Liu et~al.(2024)Liu, Feng, Xue, Wang, Wu, Lu, Zhao, Deng, Zhang, Ruan et~al.}]{liu2024deepseek}
Aixin Liu, Bei Feng, Bing Xue, Bingxuan Wang, Bochao Wu, Chengda Lu, Chenggang Zhao, Chengqi Deng, Chenyu Zhang, Chong Ruan, and 1 others. 2024.
\newblock Deepseek-v3 technical report.
\newblock \emph{arXiv preprint arXiv:2412.19437}.

\bibitem[{Meng et~al.(2024)Meng, Xia, and Chen}]{meng2024simposimplepreferenceoptimization}
Yu~Meng, Mengzhou Xia, and Danqi Chen. 2024.
\newblock \href {https://arxiv.org/abs/2405.14734} {Simpo: Simple preference optimization with a reference-free reward}.
\newblock \emph{Preprint}, arXiv:2405.14734.

\bibitem[{Minder et~al.(2025)Minder, Dumas, Juang, Chughtai, and Nanda}]{minder2025robustly}
Julian Minder, Clément Dumas, Caden Juang, Bilal Chughtai, and Neel Nanda. 2025.
\newblock Robustly identifying concepts introduced during chat fine-tuning using crosscoders.
\newblock \emph{arXiv preprint arXiv:2504.02922}.

\bibitem[{Mousi et~al.(2023)Mousi, Durrani, and Dalvi}]{mousi-etal-2023-llms}
Basel Mousi, Nadir Durrani, and Fahim Dalvi. 2023.
\newblock \href {https://doi.org/10.18653/v1/2023.emnlp-main.196} {Can {LLM}s facilitate interpretation of pre-trained language models?}
\newblock In \emph{Proceedings of the 2023 Conference on Empirical Methods in Natural Language Processing}, pages 3248--3268, Singapore. Association for Computational Linguistics.

\bibitem[{Paulo et~al.(2024)Paulo, Mallen, Juang, and Belrose}]{paulo2024automatically}
Gon{\c{c}}alo Paulo, Alex Mallen, Caden Juang, and Nora Belrose. 2024.
\newblock Automatically interpreting millions of features in large language models.
\newblock \emph{arXiv preprint arXiv:2410.13928}.

\bibitem[{Penedo et~al.(2024)Penedo, Kydl\'{\i}\v{c}ek, allal, Lozhkov, Mitchell, Raffel, Von~Werra, and Wolf}]{penedo24fineweb}
Guilherme Penedo, Hynek Kydl\'{\i}\v{c}ek, Loubna~Ben allal, Anton Lozhkov, Margaret Mitchell, Colin Raffel, Leandro Von~Werra, and Thomas Wolf. 2024.
\newblock \href {https://proceedings.neurips.cc/paper_files/paper/2024/file/370df50ccfdf8bde18f8f9c2d9151bda-Paper-Datasets_and_Benchmarks_Track.pdf} {The fineweb datasets: Decanting the web for the finest text data at scale}.
\newblock In \emph{Advances in Neural Information Processing Systems}, volume~37, pages 30811--30849. Curran Associates, Inc.

\bibitem[{Rafailov et~al.(2023)Rafailov, Sharma, Mitchell, Manning, Ermon, and Finn}]{rafailov2023direct}
Rafael Rafailov, Archit Sharma, Eric Mitchell, Christopher~D Manning, Stefano Ermon, and Chelsea Finn. 2023.
\newblock \href {https://openreview.net/forum?id=HPuSIXJaa9} {Direct preference optimization: Your language model is secretly a reward model}.
\newblock In \emph{Thirty-seventh Conference on Neural Information Processing Systems}.

\bibitem[{Singh et~al.(2025)Singh, Nan, Wang, D'Souza, Kapoor, {\"U}st{\"u}n, Koyejo, Deng, Longpre, Smith et~al.}]{singh2025leaderboard}
Shivalika Singh, Yiyang Nan, Alex Wang, Daniel D'Souza, Sayash Kapoor, Ahmet {\"U}st{\"u}n, Sanmi Koyejo, Yuntian Deng, Shayne Longpre, Noah Smith, and 1 others. 2025.
\newblock The leaderboard illusion.
\newblock \emph{arXiv preprint arXiv:2504.20879}.

\bibitem[{Team et~al.(2025)Team, Kamath, Ferret, Pathak, Vieillard, Merhej, Perrin, Matejovicova, Ram{\'e}, Rivi{\`e}re et~al.}]{team2025gemma}
Gemma Team, Aishwarya Kamath, Johan Ferret, Shreya Pathak, Nino Vieillard, Ramona Merhej, Sarah Perrin, Tatiana Matejovicova, Alexandre Ram{\'e}, Morgane Rivi{\`e}re, and 1 others. 2025.
\newblock Gemma 3 technical report.
\newblock \emph{arXiv preprint arXiv:2503.19786}.

\bibitem[{Verge(2025)}]{metascandal}
The Verge. 2025.
\newblock \href {https://www.theverge.com/meta/645012/meta-llama-4-maverick-benchmarks-gaming} {Meta got caught gaming ai benchmarks}.
\newblock Accessed: 2025-05-19.

\bibitem[{Wright and Sharkey(2024)}]{wright2024addressing}
Benjamin Wright and Lee Sharkey. 2024.
\newblock Addressing feature suppression in saes.
\newblock In \emph{AI Alignment Forum}, page~16.

\bibitem[{Xu et~al.(2024)Xu, Guan, Greene, Kechadi et~al.}]{xu2024benchmark}
Cheng Xu, Shuhao Guan, Derek Greene, M~Kechadi, and 1 others. 2024.
\newblock Benchmark data contamination of large language models: A survey.
\newblock \emph{arXiv preprint arXiv:2406.04244}.

\bibitem[{Yang et~al.(2025)Yang, Li, Yang, Zhang, Hui, Zheng, Yu, Gao, Huang, Lv et~al.}]{yang2025qwen3}
An~Yang, Anfeng Li, Baosong Yang, Beichen Zhang, Binyuan Hui, Bo~Zheng, Bowen Yu, Chang Gao, Chengen Huang, Chenxu Lv, and 1 others. 2025.
\newblock Qwen3 technical report.
\newblock \emph{arXiv preprint arXiv:2505.09388}.

\bibitem[{Yun et~al.(2021)Yun, Chen, Olshausen, and Lecun}]{yun2021transformer}
Zeyu Yun, Yubei Chen, Bruno Olshausen, and Yann Lecun. 2021.
\newblock Transformer visualization via dictionary learning: contextualized embedding as a linear superposition of transformer factors.
\newblock In \emph{Proceedings of Deep Learning Inside Out (DeeLIO): The 2nd Workshop on Knowledge Extraction and Integration for Deep Learning Architectures}, pages 1--10.

\bibitem[{Zheng et~al.(2023)Zheng, Chiang, Sheng, Li, Zhuang, Wu, Zhuang, Li, Lin, Xing et~al.}]{zheng2023lmsys}
Lianmin Zheng, Wei-Lin Chiang, Ying Sheng, Tianle Li, Siyuan Zhuang, Zhanghao Wu, Yonghao Zhuang, Zhuohan Li, Zi~Lin, Eric~P Xing, and 1 others. 2023.
\newblock Lmsys-chat-1m: A large-scale real-world llm conversation dataset.
\newblock \emph{arXiv preprint arXiv:2309.11998}.

\end{thebibliography}

\appendix

\section{LMArena Scores}
Table~\ref{tab:class-performance-default} shows the ELO scores of the two models \texttt{gemma-2-9b-it} and \texttt{gemma-2-9b-it-SimPO} with and without style correction as reported by LMArena leaderboard.
\begin{table*}[ht!]
\small
\centering
\begin{tabular}{l|rrr|rrr}
\toprule
& \multicolumn{3}{c|}{\textit{Without style control}} & \multicolumn{3}{c}{\textit{With style control}} \\
Category & \texttt{it} & \texttt{SimPO}  & Diff & \texttt{it} & \texttt{SimPO}  & Diff\\
\midrule
Math & 1177	& 1187	& 10  & 1216	& 1204	& -12\\
Instruction Following & 1177	& 1189	& 12 & 1233	& 1241	& 8\\
Coding & 1187&	1206&	19 & 1269&	1275&	6\\
Overall & 1208	&1228	& 20 &1261	&1276	& 15\\
English & 1219 & 1242&  23& 1272 & 1292&  20\\
Russian & 1198 & 1223 & 25 & 1252 & 1267 & 15 \\
Hard Prompts & 1171 &	1196	&25 & 1248 &	1265	&17\\
Multi-Turn & 1192	& 1219	& 27 & 1249	& 1269	& 20 \\
Creative Writing & 1208	& 1241	& 33 & 1256	& 1282	& 26 \\
German & 1181 & 1215 & 34 & 1236 & 1263 & 27 \\
Chinese & 1179 & 1219 & 40 & 1250 & 1277 & 27\\
\bottomrule
\end{tabular}
\caption{The LMArena Elo scores per category for \texttt{gemma-2-9b-it-SimPO} and \texttt{gemma-2-9b-it} with and without style control. The scores were extracted from the live Leaderboard on 18/9/2025.}
\label{tab:class-performance-default}
\end{table*}

\section{Latent Annotation Pipeline}
\label{appx:annotation}
Our latent categorization is semantically grounded through a structured and interpretable process. We initially experimented with general linguistic taxonomies (e.g., morphological, syntactic and discourse-level categories), but found them too narrow and ill-suited for capturing the fine-grained, capability-specific patterns emerging from the latent space. These conventional taxonomies struggle to disentangle functional concepts like hallucination detection, instruction adherence, or content moderation behaviors, which are often distributed and compositional. 

To address this, we adopted a more scalable and expressive approach using LLMs to annotate and cluster high-activation examples for each latent. These annotations are semantically rich and compositional, making them more suitable for latent concept interpretation than traditional taxonomies. This approach is grounded in and supported by prior work~\cite{mousi-etal-2023-llms, paulo2024automatically}, which demonstrates that LLMs can reliably surface latent behavioral patterns and assign meaningful labels in settings where manual annotation would be infeasible or underspecified. In this sense, we build on an emerging body of work that validates the utility of LLMs for concept-level interpretation in neural models.

 Our annotation pipeline follows a process similar to~\newcite{paulo2024automatically}. For each high-norm latent, we retrieve the top-$N$ activating documents and use  a structured prompting template (shown below, Section~\ref{appx:prompt}) to elicit a semantic description of the latent’s function. We then group, using another prompt, similar latent descriptions into one of 30 fine-grained capability categories, which were further aggregated into seven major classes (Section~\ref{appx:classes}). The latent annotation was performed using the Claude~3 Opus model (\texttt{anthropic/claude-3-opus-20240229}), which we found effective for producing consistent and interpretable concept summaries.

\subsection{Annotation prompt} 
\label{appx:prompt}
Figure~\ref{fig:prompt} shows the exact prompt used in our pipeline to annotate latents. Claude~3 Opus (\texttt{anthropic/claude-3-opus-20240229}) was provided with a set of documents that highly activate a particular latent, and asked to elicit and describe the theme or the concept of these documents. 
\begin{figure*}[t]
    \centering

\begin{Verbatim}[frame=single]
"You are an expert in neural network interpretability. I will show you several 
text examples that highly activate a specific latent (neuron/feature) in a large 
language model.

Here are the top activating documents for this latent:
Document 0:....
Document 1:....
Document N:....

Based on these examples, please:
1. Identify the common patterns, themes, concepts, or linguistic features shared
   across these documents
2. Provide a concise name/label for this latent (1-5 words)
3. Write a detailed description of what this latent appears to detect or represent
   (2-3 sentences)
4. Estimate your confidence in this interpretation (low/medium/high) and explain 
   why

Your goal is to accurately interpret what feature of language or content this 
latent is detecting."
\end{Verbatim}

    \caption{The Claude-3 Opus prompt used to categorize the latents}
    \label{fig:prompt}
\end{figure*}

\subsection{Latent categorization}
\label{appx:classes}

Table~6 shows the taxonomy acquired for SimPO latents using Claude~3 Opus (\texttt{anthropic/claude-3-opus-20240229}), comprising 7 major classes and 30 fine-grained categories.

\section{Fine-grained SimPO Results}
\label{sec:detailed-finegrained-results}
Table \ref{tab:capability-changes} shows fine-grained results for some of the top positive and negative changes between \texttt{gemma-2-9b-it} and \texttt{gemma-2-9b-it-SimPO}.

\begin{table*}[!ht]
\begin{small}
\centering
\begin{tabular}{lrrr}
\toprule
Category & IT & SimPO & Diff (\%) \\
\midrule
\multicolumn{4}{l}{\textbf{Top positive changes (SimPO > IT)}} \\
\midrule
Sexual Content Filtering & 4.46 & 7.87 & +76.2 \\
Template Following & 1.79 & 4.49 & +151.7 \\
Instruction Following & 1.79 & 4.49 & +151.7 \\
Multilingual Processing & 3.57 & 5.62 & +57.3 \\
Factual Verification & 1.79 & 3.37 & +88.8 \\
\midrule
\multicolumn{4}{l}{\textbf{Top negative changes (IT > SimPO)}} \\
\midrule
Model Self-Reference & 8.04 & 4.49 & -44.1 \\
Query Classification & 8.93 & 5.62 & -37.1 \\
Structured Output Generation & 7.14 & 4.49 & -37.1 \\
Hallucination Detection & 3.57 & 1.12 & -68.5 \\
Code Generation  & 6.25 & 4.49 & -28.1 \\
\bottomrule
\end{tabular}
\caption{Top capability changes in terms of latent counts between Gemma-2-9b-it and Gemma-2-9b-it-SimPO models}
\label{tab:capability-changes}
\end{small}
\end{table*}

\begin{table*}[t]
\small
\centering
\begin{tabular}{lrrrr}
\hline
Class & \texttt{it} & \texttt{DPO} & Diff & Change \\
\hline
Error Handling \& Quality Control &  6.25 &  8.26 &  2.01 &  +32\% \\
User Interaction Management       & 14.29 & 18.35 &  4.06 &  +28\% \\
Format \& Structure Control       & 10.71 & 12.84 &  2.13 &  +20\% \\
Safety \& Content Moderation      & 16.07 & 14.68 & -1.39 &  -9\% \\
Specialized Capabilities          & 28.57 & 25.69 & -2.88 & -10\% \\
Linguistic Capabilities           & 6.25 & 5.50 & -0.75 & -12\% \\
Information Processing            & 16.96 & 14.68 & -2.28 & -13\% \\

\hline
\end{tabular}
\caption{Comparative analysis of DPO fine-tuning effects on model behavior compared to the \texttt{it} model.}
\label{tab:dpo-vs-simpo}
\end{table*}

\section{Additional Results: DPO}
\label{appx:dpo}
To show the generality of our approach, we extend the analysis to include an investigation of Direct Preference Optimization (DPO) fine-tuning~\cite{rafailov2023direct}. The comparison between \texttt{Gemma-2-9b-it} and its DPO variant (\texttt{princeton-nlp/gemma-2-9b-it-DPO}) reveals that DPO evolves in a manner distinct from SimPO, see Table~\ref{tab:dpo-vs-simpo}, further supporting the broader applicability of our proposed approach.
In particular, our results reveal that:
\begin{itemize}
    \item DPO models exhibit improved quality control and interaction management, reflecting noticeable shifts in style, helpfulness, and politeness.
    \item However, this emphasis on stylistic alignment appears to come at the expense of safety, which receives less attention compared to SimPO.
    \item Interestingly, the process of value alignment in DPO also impacts core linguistic capabilities -- an area where SimPO demonstrates greater resilience.
\end{itemize}

These new findings strengthen our claim that our pipeline offers a replicable blueprint for systematically diagnosing model behavior changes introduced by alignment fine-tuning. They also support the value of moving beyond benchmark scores to mechanistic, latent-space-informed evaluation.

\section{Crosscoder vs. Probing}

We initially explored structural probing by applying diagnostic probes to small models ($\sim$1B parameters) using a broad suite of $\sim$100 datasets from~\cite{kantamneni2025are}. Surprisingly, even these small models performed well across many tasks, which made it difficult to detect meaningful differences via probes. This suggests that probing may lack sensitivity for surfacing nuanced differences in high-performing models, particularly when models are behaviorally similar on the surface. Furthermore, probing methods typically require prior knowledge of what to look for, which limits their effectiveness in uncovering subtle or unanticipated behavioral changes introduced by fine-tuning.

In contrast, crosscoders operate in a task-agnostic, unsupervised manner, learning to model shared and diverging latent representations between models trained on the same underlying architecture. This allows us to surface fine-grained, latent-level shifts that are often invisible through standard benchmarks or output comparisons. The technique can generalize across model pairs and training regimes without requiring carefully constructed behavioral test sets.

Moreover, recent work (e.g. \newcite{minder2025robustly}) shows that crosscoder latents can be used for causal interventions (e.g., activation patching) to test whether specific latent concepts cause behavioral changes, something that traditional probing and output comparison methods are not designed to support.

{\footnotesize
\onecolumn
\begin{longtable*}{>
{\bfseries}p{1cm} p{4cm} p{9cm}}
\multicolumn{3}{c}{Table 6: Latent taxonomy}\\
\toprule
\textbf{Code} & \textbf{Subcategory} & \textbf{Description} \\
\midrule
\endfirsthead
\toprule
\textbf{Code} & \textbf{Subcategory} & \textbf{Description} \\
\midrule
\endhead
\midrule
\multicolumn{3}{r}{{Continued on next page}} \\
\midrule
\endfoot
\bottomrule
\endlastfoot

\multicolumn{3}{l}{\textbf{A. Safety \& Content Moderation}} \\
A.1 & Harmful Content Detection & Identifies requests for violence, weapons, extremist content, or illegal activities. Activates when encountering text promoting harm or discrimination. \\
A.2 & Request Refusal Mechanisms & Recognizes when to decline inappropriate requests. Provides explanations about ethical guidelines and limitations. \\
A.3 & Jailbreak Detection & Identifies attempts to circumvent safety measures. Recognizes patterns like "evil trusted confidant" or constraint-based prompting. \\
A.4 & Sexual Content Filtering & Detects explicit sexual content requests, especially involving inappropriate scenarios. Identifies content with taboo themes or non-consensual elements. \\
A.5 & Minor Protection & Specifically focuses on protecting children in content generation. Detects requests involving minors in inappropriate contexts. \\
A.6 & Stereotype \& Bias Detection & Identifies racial, ethnic, or religious stereotyping. Detects when users request content that promotes discrimination. \\

\midrule
\multicolumn{3}{l}{\textbf{B. Linguistic Capabilities}} \\
B.7 & Multilingual Processing & Identifies non-English languages in queries. Activates language-specific response modes across multiple scripts and languages. \\
B.8 & Translation \& Language Switching & Detects requests for translation between languages. Manages language transitions within conversations. \\
B.9 & Grammar \& Style Analysis & Evaluates grammatical correctness and writing quality. Identifies spelling, syntax, and structural issues in text. \\

\midrule
\multicolumn{3}{l}{\textbf{C. Information Processing}} \\
C.10 & Summarization \& Condensing & Detects requests to summarize longer content. Extracts key information while preserving core meaning. \\
C.11 & Entity Recognition \& Extraction & Identifies specific entities (people, organizations, terms) in text. Organizes and categorizes information from unstructured content. \\
C.12 & Factual Verification & Checks consistency between summaries and source content. Verifies whether claims align with provided information. \\
C.13 & Knowledge Boundary Recognition & Identifies when information falls outside the model's knowledge. Detects when the model should acknowledge limitations rather than confabulate. \\

\midrule
\multicolumn{3}{l}{\textbf{D. User Interaction Management}} \\
D.14 & Query Classification & Categorizes types of user requests (questions, instructions, etc.). Determines appropriate response strategies. \\
D.15 & Clarification Mechanisms & Detects ambiguous or vague queries requiring additional context. Manages follow-up questioning to gather necessary information. \\
D.16 & Instruction Following & Processes and adheres to specific user instructions. Detects when constraints or formatting requirements are provided. \\
D.17 & Conversation Management & Tracks conversation history and references to previous exchanges. Maintains context across multiple turns. \\

\midrule
\multicolumn{3}{l}{\textbf{E. Format \& Structure Control}} \\
E.18 & Structured Output Generation & Formats responses as lists, tables, or other organized structures. Maintains consistent formatting patterns. \\
E.19 & JSON \& API Integration & Converts text into machine-readable formats like JSON. Structures information for downstream processing. \\
E.20 & Template Following & Detects and continues patterns established by examples. Adapts output to match specified formats. \\

\midrule
\multicolumn{3}{l}{\textbf{F. Error Handling \& Quality Control}} \\
F.21 & Self-Correction Mechanisms & Detects and acknowledges mistakes in previous responses. Provides corrections when errors are identified. \\
F.22 & Hallucination Detection & Identifies when the model is generating fabricated information. Recognizes factual inaccuracies in model outputs. \\
F.23 & Truncation Awareness & Detects when responses are about to be cut off. Identifies incomplete or abruptly ending content. \\

\midrule
\multicolumn{3}{l}{\textbf{G. Specialized Capabilities}} \\
G.24 & Code Generation \& Analysis & Produces programming code across multiple languages. Identifies errors or inconsistencies in code snippets. \\
G.25 & Professional Communication & Generates formal business content (emails, reports, etc.). Adapts tone for workplace and professional contexts. \\
G.26 & Educational Explanation & Simplifies complex topics for different knowledge levels. Provides 'Explain Like I'm 5' (ELI5) content. \\
G.27 & Creative Generation & Produces narratives, stories, and creative writing. Manages character development and dialogue. \\
G.28 & Role-Playing \& Persona Adoption & Adapts to specified character constraints. Maintains consistent persona characteristics. \\
G.29 & Text Transformation & Edits, improves, and reformats existing content. Enhances clarity and readability while preserving meaning. \\
G.30 & Model Self-Reference & Describes the model's own nature and capabilities. Manages disclosures about AI identity and limitations. \\
\end{longtable*}

\twocolumn

\end{document}